%% file: Huck2014_Arxiv_WHISPERS2014_long.tex
\documentclass[journal]{IEEEtran}
\newcommand{\argmin}[1]{\underset{#1}{\operatorname{argmin \enspace}}}
\newcommand{\norm}[1]{\left\|#1\right\|}

\ifCLASSINFOpdf
  \usepackage[pdftex]{graphicx}
  \graphicspath{{./}} 
\else
\fi
\usepackage[cmex10]{amsmath}
\usepackage{dsfont}
\usepackage{multirow}
\usepackage{amssymb}
\usepackage{graphicx}
\usepackage{algpseudocode}
\usepackage{bbm}

\usepackage[tight,footnotesize]{subfigure}
\newcommand{\R}{\mathds{R}}
\newcommand{\pscal}[2]{\left\langle#1, #2 \right \rangle}
\newcommand{\abs}[1]{\left| #1\right|}
\newcommand{\VEC}{}
\newcommand{\MAT}{\mathbf}
\newcommand{\VECS}{}
\newcommand{\MATS}{}
\newcommand{\TRANSP}{^\intercal}
\newcommand{\1}{\mathbbm 1}

\begin{document}
%
\title{
Hyperspectral pan-sharpening: \\ a variational convex constrained formulation to impose parallel level lines, solved with ADMM
}
%
%
%



\author{Alexis Huck,
        Fran\c{c}ois de Vieilleville,
        Pierre Weiss
        and~Manuel Grizonnet

\thanks{A. Huck and F. De Vieilleville are with Magellium, Toulouse, France.}
\thanks{P. Weiss is with ITAV-USR3505, universit\'e de Toulouse, France.}
\thanks{M. Grizonnet is with CNES, Toulouse, France.}}

\maketitle

\begin{abstract}
\input{abstract.tex}


\end{abstract}


\IEEEpeerreviewmaketitle


\section{Introduction}
\label{sec:introduction}
\input{introduction.tex}
\section{Problem formulation}
\label{sec:problemformulation}
\input{problemformulation.tex}
\section{ADMM based optimization}
\label{sec:algorithm}
\input{algorithm_long.tex}

\input{figures}
\section{Experimental results}
\label{sec:experiments}
\input{experiments}

\section{Conclusion}
\label{sec:conclusion}
\input{conclusion.tex}

\section*{Acknowledgment}
To cite this work, please use the reference \cite{Huck2014}.
The authors would like to thank the CNES for initializing and funding the study and providing HypXim simulated data.

\ifCLASSOPTIONcaptionsoff
  \newpage
\fi



\bibliographystyle{IEEEtran}
\bibliography{biblio_fusion}
\end{document}

%% file: abstract.tex
In this paper, we address the issue of hyperspectral \emph{pan-sharpening}, which consists in fusing a (low spatial resolution) hyperspectral image HX and a (high spatial resolution) panchromatic image P to obtain a high spatial resolution hyperspectral image. The problem is addressed under a variational convex constrained formulation. The objective favors high resolution spectral bands with level lines parallel to those of the panchromatic image. This term is balanced with a total variation term as regularizer. Fit-to-P data and fit-to-HX data constraints are effectively considered as mathematical constraints, which depend on the statistics of the data noise measurements. The developed Alternating Direction Method of Multipliers (ADMM) optimization scheme enables us to solve this problem efficiently despite the non differentiabilities and the huge number of unknowns.

%% file: introduction.tex
%
%
%
%
%

High spectral resolution of hyperspectral imaging sensors generally implies concession on spatial resolution due to optics/photonics and cost considerations.


If high spatial resolution panchromatic data are available, hyperspectral pan-sharpening can significantly help to improve the spatial resolution of sensed hyperspectral images. 


In the last three decades, pan-sharpening approaches were dedicated to multispectral data. The earliest methods were based on specific \emph{spectral-space transforms} such as the Hue-Intensity-Saturation (HIS) transform or the Principal Component Analysis (PCA) Transform. More recently, spatial frequency based approaches such as the High Pass Filter (HPF) method exploiting multiscale spatial analysis \cite{Ranchin2000} provided improved results. The multiscale spatial analysis framework generally offers very time efficient performance but lacks flexibility to consider some prior knowledge about ``physics of scene and sensor'' (the sensors Modulation Transfer Function (MTF), sensor noise or any prior information). This aspect has been a limitation for application to hyperspectral pan-sharpening. Thus, recent methods are generally based on variational \cite{Ballester2006} or bayesian \cite{Joshi2009} formulations. In particular, in \cite{Ballester2006}, the authors have proposed to consider a term based on the topographic properties of the panchromatic image. This idea stems from \cite{Caselles2000} where the authors show that most geometrical information of an optical image lies in the set of its gray level-set lines. 

The proposed algorithm includes three novelties: 1. we propose a constrained convex formulation where the constraints are the fit-to-data terms. This enables to easily tune the related parameters which are the (supposed) known noise variances of the sensors. 2. The proposed minimization algorithm is based on the ADMM. It handles the non differentiabilities, constraints and special structures of the linear transforms in an efficient way. 3. The formulation takes the MTF (Modulation Transfer Function) into account, which helps refining the fit-to-hyperspectral-data constraint. This is favorable to high spectral fidelity in pan-sharpened hypersepctral data.

%% file: problemformulation.tex
In this paper, we rearrange (hyperspectral) images into vectors in order to allow writing matrix-vector products.
Let $x =\begin{pmatrix}
                     x_1 \\ \vdots \\x_L
                    \end{pmatrix}\in \R^{LM}$ and  
$u =\begin{pmatrix}
 u_1 \\ \vdots \\u_L
\end{pmatrix}\in \R^{LN}$ denote the low spatial resolution (LR) measured hyperspectral image and the (unknown) high spatial resolution hyperspectral image respectively.
The integers $L$ and $M$ represent the number of spectral bands and the number of spectral pixels in the low resolution image, respectively. 
We let $p \in \R^N$ denote the rearranged panchromatic measured image, where $N=q^2\times M$ and $q\geq 1$ denotes the resolution factor between the low and high resolution images.
The linear projection operator which returns the $l^{th}$ spectral band is denoted $\pi_l$. 
Then, $x_l = \pi_l x \in \R^M$ and $u_l = \pi_l u \in \R^N$ are the $l^{th}$ spectral bands of $x$ and $u$ respectively. 

A model formulation for any spectral band $l$ of the hyperspectral measurements is given by
\begin{equation}
\label{eq:hypmodel}
\VEC{x}_l = \MAT D_s \MAT H_s \VEC u_l+ \VEC{n_{x_l}}.
\end{equation}
The linear operator $\MAT H_s \in \R^{N\times N}$ respresents the spatial convolution with the spatial Point Spread Function of the hyperspectral sensor. 
The linear operator $\MAT D_s \in \R^{M\times N}$ is a downsampling operator that preserves $1$ every $q$ pixels in the horizontal and vertical directions. 
Some additive sensor noise is considered in the vector $\VEC{n_{x_l}}$. 
We assume that $n_{x_l} \sim \mathcal N(0,\sigma_{x_l}^2)$ where $\sigma_{x_l}^2$ is the noise variance of the $l^{th}$ measured hyperspectral band.

A model formulation for the panchromatic image acquisition process is given by
\begin{equation}
\label{eq:panmodel}
\VEC p = \MAT G \VEC u + \VEC n_p
\end{equation}
where $\MAT G\in \R^{N \times LN}$ is a linear operator which linearly and positively combines the spectral bands with weights equal to the samples of the sensitivity spectral pattern of the panchromatic image. The noise of the measured panchromatic image is denoted $n_p \sim \mathcal N(0,\sigma_{p}^2)$.

Analogously to \cite{Ballester2006}, we will exploit the fact that the different spectral bands of hyperspectral images approximately share the same level lines. 
Such a knowledge can be integrated by comparing the gradient of the panchromatic data with the gradient of each hyperspectral image channel. 
A simple way to measure the discrepancy between the normal fields consist of using the function $f$ below
\begin{equation}
\label{eq:fdefinition}
f(u) = 
\sum_{l=1}^L
\sum_{i=1}^N
{\abs{ \pscal{\nabla  u_l(i)} {\frac{\nabla^\perp p(i)}{\norm{\nabla p(i)}_2 } }_{\R^2} } }
\end{equation}
where $\nabla = \begin{bmatrix} \MATS \partial_h \TRANSP, \MATS \partial_v \TRANSP \end{bmatrix} \TRANSP: \R^N\rightarrow \R^N\times\R^N$ is the standard discrete gradient operator, $\MATS \partial_h$ and $\MATS \partial_v$ are the horizontal and vertical gradient operators respectively, 
$\pscal{\cdot}{\cdot}_{\R^2}$ is the standard Euclidian dot product in $\R^2$ and $\norm{\cdot}_{2}$ the associated $L_2$ norm. 
The operator $\nabla^\perp=\begin{bmatrix} -\MATS \partial_v \TRANSP, \MATS \partial_h \TRANSP \end{bmatrix} \TRANSP: \R^N\rightarrow \R^N\times\R^N$ returns for each pixel a vector orthogonal to the gradient. Functional $f$ has many attractive properties: it is convex in $u$ and it can be shown to have a meaning in the continuous setting for bounded variation functions.

In natural scenes, the gradient can be very low in image areas corresponding to homogeneous radiometry of the scene. 
In such a case, $f$ does not provide much information and an additional regularizing term should be added in the variational formulation. 
In this work, we use a standard total variation regularizer \cite{Rudin1992}, commonly used for such purposes and adapted in Eq. \ref{eq:reg} to multiband images
\begin{equation}
\label{eq:reg}
TV(u) = \sum_{l=1}^L\sum_{i=1}^N\norm{(\nabla u_l)(i)}_{2}
\end{equation}
where $\norm{\cdot}_{2}$ is $L_2$-norm in $\R^2$. 
The proposed variational formulation for the hyperspectral pan-sharpening problem is as follows:
\begin{eqnarray}
\label{eq:problemformulation}
\lefteqn{\hat{u}} & = & \argmin{u}\gamma f(u)+(1-\gamma)TV(u) \\
 & s.t. & \norm{x_l- \MAT D \MAT H_s u_l}_2^2 \leq M\sigma_{x_l}^2, \forall l\in[1,\dots,L] \nonumber \\
 & & \norm{p - \MAT G u}_2^2 \leq N\sigma_p^2 \nonumber
\end{eqnarray}
where $\norm{\cdot}_2$ denotes the $L_2$ norm in $\R^M$ or $\R^N$. In this formulation, $\gamma\in [0,1]$ fixes a balance between the two terms $f$ and $TV$. The fit-to-data terms are constraints deriving from the physical models (Eq. \ref{eq:hypmodel} and \ref{eq:panmodel}). The parameters $\{\sigma_{x_l}\}_{l\in\{1,\dots,L\}}$ and $\sigma_p$ can be \emph{a priori} given or estimated, which is a strong asset of the variational constrained formulation. 

%% file: algorithm_long.tex
%

The proposed algorithm is called TVLCSP (for Total Variation iso-gray Level-set Curves Spectral Pattern) and its pseudo-code is given in the procedure TVLCSP. 
A variant (called TVLC) not considering the sensitivity spectral and fit-to-panchromatic data has been developed but is not presented here.

 \begin{figure}
 \begin{algorithmic}[1]
 \Procedure{TVLCSP}{$\VEC x, \VEC p,\sigma_x, \sigma_p, \alpha, \beta$ }
 \State \# Initialization
 \State $\VECS \lambda \gets 0$ \Comment{a vector of zeroes.}
 \State $\VEC y_1 \gets \left[ \uparrow \VEC x \TRANSP, \uparrow \VEC x \TRANSP \right] \TRANSP$, 
 $\VEC y_2 \gets \left[ \uparrow \VEC x \TRANSP, \uparrow \VEC x \TRANSP \right] \TRANSP$, 
 \State $\VEC y_3 \gets \uparrow \VEC x$, $\VEC y_4 \gets \VEC p$
 
 \State \# Iterative scheme
 \While{stop condition not met}
 
 \ForAll{$l\in\{1,\dots,L\}$}
 \State
 $\VEC z_{1,l} \gets \nabla \VEC u^l - \frac{ \VECS \lambda_1^l}{\beta}$
 \State
 $\VEC y_1^l \gets \frac{\VEC z_1^l}{\norm{\VEC z_1^l}_{2,\R^2}} \cdot \max\left( \norm{\VEC z_1^l}_{2,\R^2} - \frac{\gamma}{\beta}, 0 \right)$
 
 \State $\epsilon \gets \pscal{\VEC y_2^l}{\VECS \eta}_{\R^2}$ \Comment{where $\eta = \frac{ \nabla^\perp p } {\norm{\nabla p}_2 }$}
 
 \If{$\epsilon \neq \VEC 0$}
 \State $\VEC y_2^l \gets \nabla \VEC u^l - \frac{1}{\beta}\left( \VECS \lambda_2^l + (1- \gamma) \mbox{sign}(\epsilon) \VECS \eta \right)$
 \Else
 \State
 $\VEC z_2^l \gets \nabla \VEC u - \frac{\VECS \lambda_2^l}{\beta}$
 \State
 $\VECS \alpha \gets \beta \frac{\pscal{\VEC z_2^l}{\VECS \eta}_{\R^2}}{(1-\gamma) \norm{\VECS \eta}_{2,\R^2}^2}$
 \State
 $\VEC y_2^l \gets \left( \1 - \frac{1-\gamma}{\beta}\VECS \alpha\right) \cdot \VEC z_2^l$
 \EndIf
 
 \State $\VEC z_3^l \gets \frac{\VECS \lambda_3^l}{\beta} - \MAT H_s \VEC u$
 \If{$\norm{\VEC x^l - \MAT D_s \MAT H_s \VEC u^l}_2^2 \leq M \sigma_x^2$}
 \State $\VEC y_3^l \gets \VEC z_3^l$
 \Else
 \State $\delta \gets \frac{\norm{\VEC x^l - \MAT D_s \MAT H_s \VEC u^l}_2 - \sqrt{M}\sigma_x}{\sqrt{M}\sigma_x}$
 \State $\VEC y_3^l \gets \left[ \mbox{Id} - \frac{\delta}{\beta+\delta} \MAT D_s \TRANSP \MAT D_s \right] \VEC z_3^l + \frac{\delta}{\beta+\delta} \MAT D_s \TRANSP \VEC x$
 \EndIf
 
 \EndFor
 
 \State $\VEC z_4 \gets \frac{\VECS \lambda_4}{\beta} - \MAT H_\lambda \VEC u$
 \If{$\norm{\VEC p - \MAT D_\lambda \MAT H_\lambda \VEC u} _2^2\leq N \sigma_p^2$}
 \State $\VEC y_4 \gets \VEC z_4$
 \Else
 \State $\delta \gets \frac{\norm{\VEC p - \MAT D_\lambda \MAT H_\lambda \VEC u}_2 - \sqrt{N}\sigma_p}{\sqrt{N}\sigma_p}$
 \State $\VEC y_4 \gets \left[ \mbox{Id} - \frac{\delta}{\beta+\delta} \MAT D_\lambda \TRANSP \MAT D_\lambda \right] \VEC z_4 + \frac{\delta}{\beta+\delta} \MAT D_\lambda \TRANSP \VEC p$
 \EndIf
 \State $\VEC u \gets \left( \MAT M \TRANSP \MAT M \right)^{-1} \MAT M \TRANSP \left( \frac{\VECS \lambda}{\beta} + \VEC y \right)$ \label{line:u}
 \State $\lambda \gets \lambda + \beta \left( y - \MAT M u\right)$
 \EndWhile
 
 \State \textbf{return} $u$
 
 \EndProcedure
 \end{algorithmic}
 \end{figure}

 In the procedure, the vectors $\1$ and matrices Id are vectors of ones and identity matrices. Their dimensions depend on the context.
 $\forall k \in \{1,2,3\}, \VEC y_k=\left[ \VEC y_{k,1} \TRANSP,\dots,\VEC y_{k,L} \TRANSP \right] \TRANSP$ and $ \VECS \lambda_k=\left[ \VECS \lambda_{k,1} \TRANSP,\dots,\VECS\lambda_{k,L} \TRANSP \right] \TRANSP$ such that $\forall l \in \{1,\dots,L\}, \forall k\in\{1,2\}, \VEC y_{k,l}, \VECS \lambda_{k,l} \in \R^N\times \R^N$ and $\VEC y_{3,l} \in \R^N$. We define $\VECS \lambda = \left[ \VECS \lambda_{1} \TRANSP,\dots,\VECS \lambda_4 \TRANSP \right] \TRANSP$.
 We can note that the operator $\MAT G: \R^{LN} \rightarrow \R^N$ can be decomposed as:
 \begin{equation*}
 \MAT G = \MAT D_\lambda \MAT H_\lambda
 \end{equation*}
 where $\MAT H_\lambda : \R^{LN} \rightarrow \R^{LN}$ is a circulant matrix associated with a spatially invariant convolution kernel defined by the sensitivity spectral pattern of the panchromatic image and $\MAT D_\lambda:\R^{LN} \rightarrow \R^N$ is a spectral decimation operator.
 
 
 Finally, we define the matrix
%
 \begin{equation}
 \label{eq:Mdefinition} 
 \MAT M = \left[ ( \nabla \pi_1)\TRANSP, \dots, (\nabla \pi_L)\TRANSP, (\nabla \pi_1)\TRANSP, \dots, (\nabla \pi_L)\TRANSP,
 \MAT H_s \TRANSP , \MAT H_\lambda \TRANSP
 \right] \TRANSP
 \end{equation}

The up-sampling operator $\uparrow: \R^{LM} \rightarrow \R^{LN}$ spatially up-samples a vectorized hyperspectral image by a factor $q$ and the vector $\1$ is a vector of ones of suitable dimension.
The multiplications and divisions are element-wise. The ADMM procedure introduces an internal parameter, denoted $\beta$, whose value impacts the convergence speed.
 
Note that the update rule of $\VEC u$ (line $34$ in the procedure) can be computed in the Fourier domain. More precisely, since $\MAT M \TRANSP$ and $\MAT M \TRANSP \MAT M$ are circulant matrices (as concatenation and summation of circulant matrices, respectively), the left product by $\MAT M \TRANSP$, the inversion of $\MAT M \TRANSP \MAT M$ and the left product by $\left(\MAT M \TRANSP \MAT M\right)^{-1}$ can be performed in the Fourier domain.

matrix $\MAT M$ is a concatenation of circulant matrices (Eq. \ref{eq:Mdefinition}), each associated with a convolution-type operation either in the two spatial dimensions (case of all matrices in Eq. \ref{eq:Mdefinition} but $\MAT H_\lambda \TRANSP$) or in the spectral dimension (case of $\MAT H_\lambda \TRANSP$). Thus, the left side product by $\MAT M \TRANSP$ is in the Fourier domain. Additionally, $\MAT M \TRANSP \MAT M$ is a summation of circulant matrices, which is also circulant so the left-product by its inverse can be computed in the Fourier domain too. Thus, line $34$ of the procedure should be replaced by:

 \begin{equation}
 \label{eq:ufourier}
 u=\mathcal F^{-1} \left( \frac{\mathcal F\left( \MAT M \TRANSP \right) \cdot \mathcal F\left( \frac{\VECS \lambda}{\beta} + \VEC y \right)}
 {\mathcal F \left( \MAT M \TRANSP \MAT M\right)} \right)
 \end{equation}
 where $\mathcal F$ and $\mathcal F^{-1}$ represent the Fourier transform and its inverse, respectively.
 
 Currently, the stop condition is a number of iterations but another approach could be based on the stationary of $y$, $\lambda$ and $u$.

%% file: figures.tex

\begin{figure}[ht!]
\centering
\subfigure[HX LR]{\includegraphics[width=0.45\columnwidth]{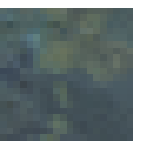}}
\subfigure[Panchromatic HR]{\includegraphics[width=0.45\columnwidth]{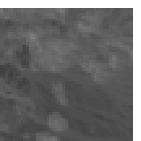}}
\subfigure[Wavelet HR estimation]{\includegraphics[width=0.45\columnwidth]{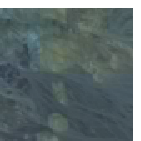}}
\subfigure[TVLCSP HR estimation]{\includegraphics[width=0.45\columnwidth]{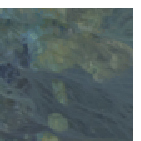}}

\caption{Cuprite scene and processing with wavelet and TVLCSP, for a resolution ratio $q=4$}
\label{fig:cuprite}
\end{figure}



%
\begin{figure}
\centering
\subfigure[Panchromatic spectral pattern]{\includegraphics[width=0.45\columnwidth]{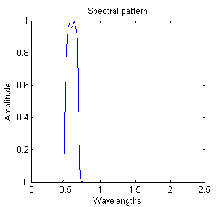}}
\subfigure[Panchro]{\includegraphics[width=0.45\columnwidth]{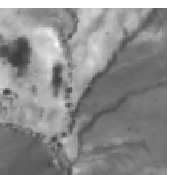}}
\subfigure[LR HSI - q=2]{\includegraphics[width=0.45\columnwidth]{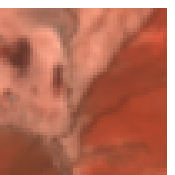}}
\subfigure[TVLCSP - q=2]{\includegraphics[width=0.45\columnwidth]{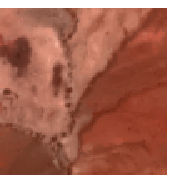}}
\subfigure[LR HSI - q=4]{\includegraphics[width=0.45\columnwidth]{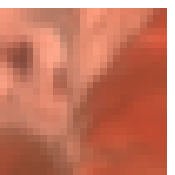}}
\subfigure[TVLCSP - q=4]{\includegraphics[width=0.45\columnwidth]{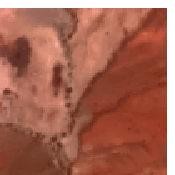}}
\subfigure[LR HSI - q=6]{\includegraphics[width=0.45\columnwidth]{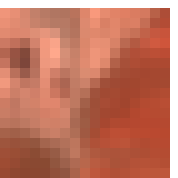}}
\subfigure[TVLCSP - q=6]{\includegraphics[width=0.45\columnwidth]{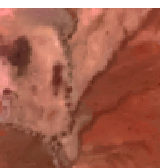}}
\caption{HypXim scene and processing with TVLCSP, for resolution ratios $q=2$ and $q=6$.}
\label{fig:hypxim_sc2}
\end{figure}

%% file: experiments.tex
We present here results of TVLCSP on AVIRIS \cite{Aviris} and simulated HypXim \cite{Marion2011} data.
We have first extracted a selection of the Cuprite scene (AVIRIS) which represents a mineral area. The $224-$spectral band data has been preprocessed and simulated as follows. 1 - Absorption spectral bands have been removed (bands: $1-6$, $106-114$, $152-170$ and $215-224$) to get a reference high resolution hyperspectral image $u_{\text{ref}}$. 2 - A convex combination of the spectral bands of $u_{\text{ref}}$ gives the simulated panchromatic data $p$. The weights are the coefficients of the vector $\mathbf g = \left[\frac{1}{80}\dots\frac{1}{80}\right]$. 3 - The low resolution hyperspectral image $x$ has been obtained from Eq. \ref{eq:hypmodel} without noise and with $\MAT H_s$ representing an average filter.

The chosen algorithm parameters are given in Table \ref{table:aviris}.
\begin{table}
\centering
\begin{scriptsize}
\begin{tabular}{|c|c|c|c|c|c|}
\hline
q 	& $\beta$ 	& $\gamma$ 	& $\sigma_p$ 		& $\sigma_{\tilde{\mathbf u}}$	& \#iter			\\
\hline
$4$ 	& $1000$ 			& $0.01$ & 		$0.0001$ 					& $0.0001$		& $300$ \& $3000$	\\
\hline
\end{tabular}
\caption{Parameters of TVLCSP for the tests on the Cuprite scene.}
\label{table:aviris}
\end{scriptsize}
\end{table}
Visual results are presented in Fig. \ref{fig:cuprite}.

In Table \ref{table:cuprite} we present quantitative evaluation and comparison with a wavelet-based pan-sharpening method \cite{Ranchin2000} using usual performance metrics: 1 - global quality metrics RMSE and ERGAS, 2 - spectral quality metrics SAM and the spectral dispersion the spatial dispersion $D_{\lambda}$ \cite{Alparone2008}), and 3 - spatial quality metrics FCC \cite{Zhou1998} and spatial dispersion $D_{s}$ \cite{Alparone2008}. Note that $D_{\lambda}$ and $D_s$ are metrics without reference (ground truth high resolution hyperspectral image) requirement, which is relevant where no reference is available or when the reference is likely to introduce error in comparison (case of our HypXim data) due to noise.

%
\begin{table}
\centering
\begin{scriptsize}
\begin{tabular}{|l|c|c|c|c|}
\cline{2-4}
\multicolumn{1}{c|}{}   	& \multicolumn{2}{c|}{TVLCSP}	& \multirow{2}{*}{Wavelet} \\ 
\cline{2-3}
\multicolumn{1}{c|}{}			& \#300    				& \#3000		&         \\
\hline
RMSE ($\times100$) 				& \textbf{0.48} 	& 0.59 			& 0.91 		\\
ERGAS               			& \textbf{5.45}		& 6.68 			& 10.3 		\\
SAM                 			& \textbf{0.61} 	& 0.70 			& 0.88 		\\
FCC ($\times100$)  				& \textbf{99.3} 	& 99.0 			& 99.1 		\\
$D_s$ ($\times100$) 			& 1.15 						& \textbf{0.89} 			& 2.35 		\\
$D_\lambda$ ($\times100$) & 1.82 						& \textbf{1.22} 			& 4.43 		\\
\hline
\end{tabular}
\caption{Performances of pan-sharpening algorithms on the Cuprite subimage.}
\label{table:cuprite}
\end{scriptsize}
\end{table}

Additionally, TVLCSP has been tested on simulated HypXim data. They have been simulated from data acquired in the framework of the Pl\'eiades program. The scene is located in Namibia and a sub-scene has been extracted. Some characteristics of the considered data are given in Table \ref{table:hypximdata}.
\begin{table}
\centering
\begin{scriptsize}
\begin{tabular}{|c|c|c|}
\hline
\multirow{2}{*}{$q$} 		& Spatial			& \multirow{2}{*}{Sumulated sensor}		\\
					& resolution (m)	& 								\\
\hline
$1$					& $4.80$			& Panchromatic sensor				\\
$1$					& $4.80$			& Reference						\\
$2$					& $9.60$			& HypXim P (Performance concept) 	 	\\
$4$ 					& $19.20$			& HypXim C (Challenging concept) 		\\
$6$					& $28.80$			& ENMAP 					 	\\
\hline
\end{tabular}
\caption{Characteristics of the simulated HypXim and panchromatic data.}
\label{table:hypximdata}
\end{scriptsize}
\end{table}
%
The considered sensitivity spectral pattern is shown in Fig. \ref{fig:hypxim_sc2}(a). We see that only some ($20$) of the spectral bands contribute to the panchromatic data thus only $20$ non-zero coefficients in $\mathbf g$ and the presented results only concerns these bands. Note that the hyperspectral sensor spatial Point Spread Functions (PSF) has been supposed Gaussian spectrally and spatially invariant, with a parameter tuned experimentally. The visual results are presented in Fig. \ref{fig:hypxim_sc2} and the corresponding performance metrics are given in Table \ref{table:brgm-bandcase1-sc2}.
\begin{table}
\centering
\begin{scriptsize}
\begin{tabular}{|l|c|c|c|c|c|c|}
\cline{2-4}
\multicolumn{1}{c|}{}			& $q=2$			& $q=4$     			& $q=6$   		\\
\hline
RMSE $\times 100$ 			& 1.73			& 2.08				& 2.29				\\
ERGAS               			& 11.5			& 28.5				& 47.0				\\
SAM                 				& 1.57			& 1.50				& 1.64				\\
FCC $\times 100$   			& 97.6			& 97.3				& 97.2				\\
$D_s$ $\times 100$			& 3.05	 		& 4.28				& 4.59				\\
$D_\lambda$ $\times 100$ 	& 3.41			& 2.30 				& 1.82				\\
\hline
\end{tabular}
\caption{Performances of TVLCSP on the HypXim sub-image}.
\label{table:brgm-bandcase1-sc2}
\end{scriptsize}
\end{table}
We see that the method works well for many resolution rations. However, the results on HypXim are not as good as those on AVIRIS, probably due to our approximation hypotheses on the sensor parameters and to the presence of noise. Note that the simulated reference image is corrupted by sensor noise whereas TVLCSP provides relatively denoised data estimations, which introduces lack of confidence in the performance metrics values.

%% file: conclusion.tex
We have tackled the pan-sharpening problem using a variational convex constrained approach with an objective based on the conservation of the set of iso-gray-level lines among spectral bands and total variation. The fit-to-data constraints have been mathematically considered as such and are based on the signal model and the sensor parameters, including noise statistics. An ADMM scheme has been developped, called TVLCSP and evaluated on AVIRIS and HypXim simulated data.